\documentclass[sigconf]{acmart}

\acmConference[]{}{2025}{}

\usepackage{graphicx} %
\usepackage{array} 
\usepackage{multirow}
\usepackage{longtable}
\usepackage{booktabs}
\usepackage{geometry}
\usepackage{graphicx}
\usepackage{caption}
\usepackage{subcaption}
\usepackage{enumitem}
\usepackage{float}
\usepackage{placeins}
\usepackage{makecell}
\usepackage{stfloats}
\usepackage{pifont}
\usepackage{tabularx}

\newcommand{\fm}{\textit{FM-FoG}}

\AtBeginDocument{%
  }

\begin{document}
\title[\fm]{\fm: A Real-Time Foundation Model-based Wearable System for Freezing-of-Gait Mitigation}

\author{Chuntian Chi}
\affiliation{%
  \institution{Department of Computer Science, William \& Mary}
  \city{Virginia}
  \country{USA}}
\email{cchi05@wm.edu}

\author{John Clapham}
\affiliation{%
  \institution{Department of Computer Science, William \& Mary}
  \city{Virginia}
  \country{USA}}
\email{jmclapham@wm.edu}

\author{Leslie Cloud}
\affiliation{%
  \institution{Department of Neurology, \\ Virginia Commonwealth University}
  \city{Virginia}
  \country{USA}}
\email{leslie.cloud@vcuhealth.org}

\author{Ingrid Pretzer-Aboff}
\affiliation{%
  \institution{School of Nursing, \\ Virginia Commonwealth University}
  \city{Virginia}
  \country{USA}}
\email{iaboff@vcu.edu}

\author{GinaMari Blackwell}
\affiliation{%
  \institution{Department of Neurology, Virginia Commonwealth University}
  \city{Virginia}
  \country{USA}}
\email{gina.blackwell@vcuhealth.org}

\author{Huajie Shao}
\affiliation{%
  \institution{Department of Computer Science, William \& Mary}
  \city{Virginia}
  \country{USA}}
\email{hshao@wm.edu}

\author{Gang Zhou}
\affiliation{%
  \institution{Department of Computer Science, William \& Mary}
  \city{Virginia}
  \country{USA}}
\email{gzhou@wm.edu}

\renewcommand{\shortauthors}{Chi et al.}

\begin{abstract}
Freezing-of-Gait (FoG) affects over 50\% of mid-to-late stage Parkinson's disease (PD) patients, significantly impairing patients' mobility independence and reducing quality of life. FoG is characterized by sudden episodes where walking cannot start or is interrupted, occurring exclusively during standing or walking -- never while sitting or lying down. Current FoG detection systems require extensive patient-specific training data and lack generalization, limiting clinical deployment. To address these issues, we introduce FM-FoG, a real-time foundation model-based wearable system achieving FoG detection in unseen patients without patient-specific training. Our approach combines self-supervised pretraining on diverse Inertial Measurement Unit (IMU) datasets with sensor context integration. Since FoG occurs only during ambulatory activities, a lightweight CNN-LSTM activity classifier selectively activates the foundation model only during walking or standing, avoiding unnecessary computation. Evaluated on the VCU FoG-IMU dataset with 23 PD patients, FM-FoG achieves a 98.5\% F1-score when tested on previously unseen patients, substantially outperforming competitive baseline methods. Deployed on a Google Pixel 8a smartphone, the system extends battery life by up to 72\% while maintaining sub-20ms intervention latency. The results indicate that our FM-FoG 
can enable practical, energy-efficient healthcare applications that generalize across patients without individual training requirements.
\end{abstract}




\maketitle

{
\section{Introduction}

\setlength{\textfloatsep}{6.294pt plus 1pt minus 1pt}
Parkinson's disease (PD) affects approximately 10 million people worldwide, with prevalence increasing dramatically with age \cite{GBD2016ParkinsonsDiseaseCollaborators2018}. Freezing-of-Gait (FoG) is one of the most debilitating motor symptoms of PD, affecting over 50\% of individuals with mid-to-late stage PD \cite{Nutt2011FreezingOfGait}. FoG is characterized by sudden episodes where walking cannot start or is interrupted despite the effort to move forward, creating a profound loss of movement independence. Beyond its direct impact on gait function, FoG significantly increases fall risk and hospitalizations \cite{Bloem2004FallsAndFreezing, Wielinski2005FallsAndInjuries}. This combination of mobility impairment and safety concerns makes FoG detection and management a critical healthcare priority affecting millions of patients around the world \cite{DorseyBloem2018ParkinsonPandemic}. 

Current research approaches for FoG management primarily rely on detection systems designed to identify freezing episodes as they occur, enabling timely intervention through cueing techniques that help patients resume normal walking \cite{maziluGaitAssist, Delval2014AuditoryCueingOfGait}. However, existing FoG detection methods suffer from several critical \textit{limitations} that impede their practical clinical utility: (1) Dependence on patient-specific training data, requiring extensive, costly, and time-consuming data collection procedures for each individual \cite{BORZI2023102459, Bachlin2010WearableAssistant, Pham2017FreezingOfGaitDetection}; (2) Poor generalization across subjects, reflecting the high inter-subject variability in FoG manifestation patterns \cite{djuricFoGdetection, Tripoliti2013AutomaticDetectionFreezing}; (3) Reliance on supervised learning with manual labeling, which is exhaustive, prone to inconsistency, and subject to expert disagreement \cite{Palmerini2017Identification, RodriguezMartin2017HomeDetectionFreezing, Koltermann2023FoGFinder,Gait-Guard, Chen2025SoTA}; (4) Dependence on domain-specific feature engineering, limiting their adaptability across different sensor configurations and deployment scenarios \cite{SAMA2018135,Gait-Guard,Moore2008AmbulatoryMonitoring,Zach2015IdentifyingFreezing}; (5) Continuous model inference requirements, leading to inefficient battery usage and making them impractical for real-world mobile deployment \cite{Moore2008AmbulatoryMonitoring, RodriguezMartin2017HomeDetectionFreezing}.


\begin{table}[!htbp]
  \centering
  \caption{Sensor placement locations across datasets}
  \label{tab:datasets-sensors}
  \resizebox{\linewidth}{!}{
  \begin{tabular}{|l|c|c|c|c|c|c|}
    \hline
    \bfseries Dataset    & \bfseries Ankle & \bfseries Wrist & \bfseries Chest & \bfseries Thigh & \bfseries Trunk & \bfseries Lower Back \\
    \hline
    tDCS FOG & & & & & & \(\checkmark\) \\
    \hline
    DeFOG & & & & & & \(\checkmark\) \\
    \hline
    Daphnet & \(\checkmark\) & & & \(\checkmark\)& \(\checkmark\)& \\
    \hline
    PAMAP2 & \(\checkmark\) & \(\checkmark\) & \(\checkmark\) & & & \\
    \hline
    Ours & \(\checkmark\) & & & & & \\
    \hline
  \end{tabular}
  }
\end{table}

Inspired by recent success of large language models (LLM), we propose to develop foundation models to enhance domain generalization and reduce human annotated data.
However, directly applying foundation model principles to FoG detection faces significant challenges due to the unique characteristics of IMU sensor data. As shown in Table \ref{tab:datasets-sensors}, 
wearable IMU sensors are deployed across diverse body locations. This heterogeneity creates fundamental challenges: (1) Spatial heterogeneity: sensors at different locations (ankle, wrist, chest, etc.) capture unique movement patterns for the same underlying motor activity, necessitating spatial context awareness; (2) Temporal heterogeneity: datasets collect data at varying temporal resolutions (64Hz, 100Hz, 128Hz), creating irregularly sampled time-series data. Unlike conventional transformers operating on fixed-resolution textual or visual tokens, IMU-based approaches must handle these real-world data challenges. The spatial and temporal heterogeneity necessitates context-aware integration into existing FMs. 

These challenges raise three fundamental research questions:

\begin{enumerate}
    \item \textbf{RQ1:} How can we largely eliminate the need for patient-specific data collection and annotation?
    \item \textbf{RQ2:} Within this foundation model framework, how can we effectively learn representations from IMU data collected under varying conditions (sensor placement locations and sampling frequencies) to enable FoG detection generalizable to unseen patients?
    \item \textbf{RQ3:} How can we deploy such a foundation model with minimal energy consumption to enable real-time FoG management in practical wearable systems?
\end{enumerate}

\noindent
\textbf{Our Contributions.} 
For RQ1, we develop FM-FoG, an IMU foundation model specifically designed for FoG detection that can generalize across patients. Our approach demonstrates that foundation model principles can be successfully adapted to wearable sensor data in healthcare applications and eliminates the need for extensive patient-specific training data.

For RQ2, we address the challenge of spatial data heterogeneity by incorporating context into the IMU foundation model; and we harmonize irregularly sampled data using cubic-spline interpolation techniques for temporal alignment.

For RQ3, we design and implement the first energy-efficient event-triggered foundation model that strategically activates the foundation model only when needed. This approach achieves a substantial reduction in energy consumption compared to continuous monitoring approaches while maintaining comparable detection performance.

Overall, we propose the first context-aware foundation model for real-time FoG detection in new patients without individual training. FM-FoG demonstrates superior FoG detection performance, achieving a 98.5\% F1-score while maintaining practical deployment feasibility with 19.9ms intervention latency and extending battery life by up to 72\% compared to continuous monitoring approaches through event-triggered activation. The system operates efficiently on smartphones with only 1.2M parameters, consuming 16.1\% CPU and 112MB memory, making it suitable for real-world deployment. 

The remainder of this paper is structured as follows. Section 2 reviews related work in FoG management with classical ML methods, foundation models, and unsupervised learning for healthcare. Section 3 provides an overview of our proposed system. Section 4 details our methodology, including our datasets, the self-supervised pretrain-then-fine-tune approach for training FM-FoG, and the event-triggering activity classifier. Section 5 presents a comprehensive experimental evaluation that demonstrates the FoG detection performance with cross-patient generalization and system performance analysis. Section 6 discusses limitations and future research directions. Section 7 concludes this paper.
\section{Related Work}
In this section, we review prior work on FoG management, the use of foundation models in healthcare and Internet of Things (IoT), and related unsupervised learning approaches. We highlight their strengths and limitations to position our proposed FM-FoG system.
\subsection{FoG Management with Classical ML}
\textbf{IMU-based FoG Detection with ML.}
IMU sensors are widely used for FoG detection due to their portability and non-invasive nature. Notable systems include FoG-Finder \cite{Koltermann2023FoGFinder} using multi-input Convolutional Neural Networks (CNNs) with frequency-domain analysis, Gait-Guard \cite{Gait-Guard} with turn-aware transformer neural networks, and the dual-level approach \cite{Chen2025SoTA} combining out-of-distribution detection with anomaly detection. These build upon earlier work by Mazilu et al. \cite{MAZILU20161} with wrist sensors, Pham et al. \cite{Pham2017FreezingOfGaitDetection} with anomaly scores, DeepFoG \cite{Bikias2021DeepFoG}, wavelet transform methods \cite{RezvanianLockhart2016RealTimeDetectionFoG}, multi-head CNNs \cite{BORZI2023102459}, and recurrent neural network approaches \cite{wangFoG}. However, all approaches require supervised pretraining, patient-specific fine-tuning, and lack energy optimization for real-world deployment.

Our approach addresses these limitations by leveraging foundation models that learn generalizable representations from diverse sensor data without requiring patient-specific training, while maintaining energy efficiency.


\noindent\textbf{FoG Detection with Other Sensor Modalities.} Alternative sensing modalities include plantar pressure sensors \cite{Marcante2021FootPressureWearableSensors,Shalin2021PredictionAndDetection,Pardoel2022PredictionUnilateralBilateral}, smart shoes \cite{Patil2022FreezeFallDetection,Prado2021ContinuousIdentification}, vision-based approaches using graph representations \cite{Hu2020VisionBasedFoG} and graph fusion networks \cite{Hu2023GraphFusionNetwork}, motion cue analysis \cite{Khan2013MotionCueAnalysis}, and Red-Green-Blue camera-based gait classification \cite{NietoHidalgo2016VisionBasedProposal}.

Additional modalities include EEG-based prediction \cite{Handojoseno2014AnalysisPredictionFOG,Handojoseno2018PredictionOfFOG}, multimodal sensor fusion \cite{Bajpai2023MultimodalModelFusion}, and contactless methods using WiFi and radar imaging \cite{Shah2020SensorFusion, Tahir2019WiFreeze} coupled with deep learning techniques.

While promising, these alternative modalities suffer from privacy concerns (vision-based), limited practicality for daily use (EEG), reduced portability (foot pressure systems), or restricted mobility (WiFi/radar systems) compared to IMU-based approaches.

\subsection{Foundation Models}
\textbf{Foundation Models for Healthcare.} FMs have emerged as a powerful AI tool in healthcare, enabling knowledge transfer across diverse medical tasks. Moor et al. present a comprehensive overview of these general medical AI systems \cite{Moor2023FoundationModelsMedicalAI}, while Wornow et al. examine FMs applications to electronic health records \cite{Wornow2023ShakyFoundationsEMR}, highlighting both promise and challenges. Recent innovations integrate wearable sensor data with FMs, with Kim et al. showing LLMs can enhance health prediction \cite{Kim2024HealthLLM}, while Liu et al. explored few-shot learning capabilities for health monitoring \cite{Liu2023FewShotHealthLLM}. Yang et al. \cite{Yang2024DrHouse} and Fang et al. \cite{Fang2024PhysioLLM} introduced LLM-empowered systems for diagnostic reasoning by combining sensor data with expert knowledge. 

Despite their potential, healthcare FMs face significant deployment challenges due to computational requirements. While excelling at general medical knowledge representations, they typically lack specialized capabilities for specific symptom monitoring. Most models struggle with continuous sensor data streams essential for movement disorder applications, limiting their effectiveness for tasks such as FoG detection. Notably, none of the existing approaches have developed FMs specifically tailored for IMU data.

\vspace{1.5em}

\noindent\textbf{Foundation Models for IoT.} FMs have also advanced IoT applications. Yang et al. introduced EdgeFM for open-set learning on resource-constrained edge devices \cite{Yang2024EdgeFM}, while Baris et al. analyzed the unique challenges of CPS-IoT systems \cite{Baris2025FoundationModelsCPSIoT}. Several works have focused on IoT sensing capabilities, including Xue et al.'s zero-shot IoT approach \cite{Xue2024ZeroShotIoTSensing}, Kimura et al.'s VibroFM for multimodal sensing \cite{Kimura2024VibroFM}, and Kara et al.'s self-supervised learning framework \cite{Kara2024PhyMask}. An et al. explored integrating LLMs with IoT systems for enhanced task reasoning \cite{An2024IoTLLM}, and Han et al. investigated complex event detection in CPS-IoT environments \cite{Han2025OnlineCEDCPSIoT}.

IoT FMs face several limitations, including difficulties handling irregular sampling rates and diverse sensor configurations typical in real-world deployments. Edge computing constraints require model compression that often compromises performance. Current approaches mostly focus on basic perception tasks rather than specific downstream applications, and the heterogeneity of IoT data sources complicates the development of generalizable models. Importantly, while several works address general IoT sensing, to the best of our knowledge, no prior work has developed a foundation model specifically designed for IMU data and its unique characteristics relevant to specific healthcare applications such as FoG detection.

\subsection{Unsupervised Learning for Healthcare}
Unsupervised learning approaches offer valuable tools for healthcare applications when labeled data is scarce. Khowaja et al. addressed data heterogeneity in medical images through self-supervised federated learning \cite{Khowaja2025SelfFed}, and Wang et al. discovered latent disease clusters in electronic health records using unsupervised techniques \cite{Wang2020UnsupervisedMachineLearning}. For anomaly detection, Zimmerer et al. \cite{Zimmerer2018ContextEncodingVAE} and Bijlani et al. \cite{Bijlani2022GCMP} developed unsupervised methods for medical images and sensor-based health monitoring, respectively. Liu et al. \cite{Liu2021AutomatedCardiacSegmentation} tackled cross-modal medical image analysis using unsupervised multi-domain adaptation.

Most relevant to our FoG detection is ADMarker by Ouyang et al. \cite{Ouyang2024ADMarker}, a multi-modal federated learning system for monitoring digital biomarkers of Alzheimer's disease that utilized IMU sensor data. 

Current unsupervised approaches in healthcare mostly focus on image and text data, which lack the ability to handle irregular sampling rates and varying sensor placements unique to IMU data. These systems also lack energy efficiency considerations, limiting their deployment on wearable devices. Most critically, they generally do not support generalization to unseen subjects, requiring additional model adaptation or retraining for each new user.
This requirement, combined with limited transferability, restricts their application to heterogeneous neurological conditions like PD. For PD patients, symptom manifestation varies dramatically across individuals and disease stages, even among those at similar progression levels. This variability demands adaptive continuous monitoring solutions capable of generalization.
\section{System Overview}
\begin{figure}[!tb]
    \centering
    \includegraphics[width=\linewidth, trim=13.5cm 3.5cm 13cm 2cm, clip]{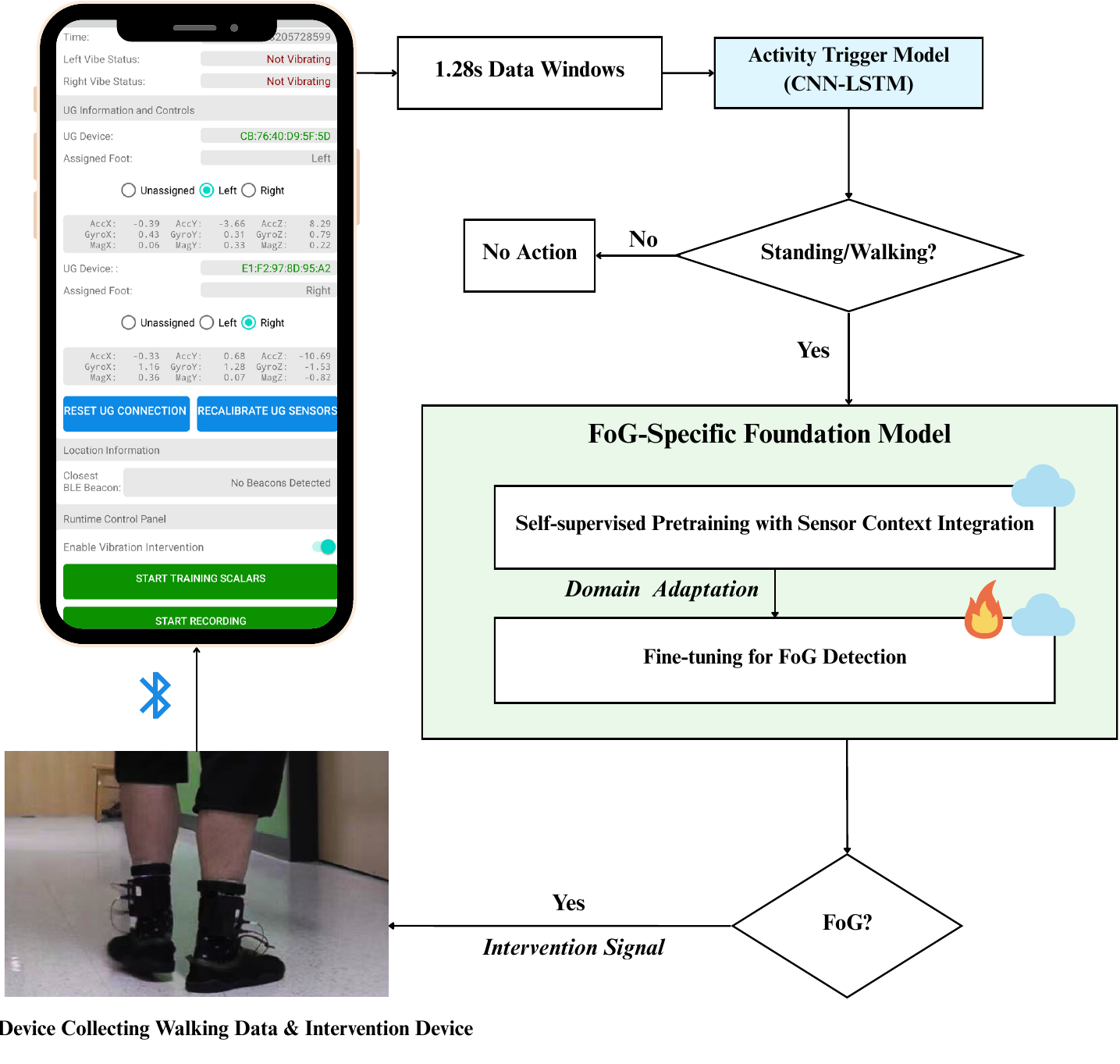}
    \caption{FM-FoG System Overview. It consists of two main components: an activity trigger model (CNN-LSTM), and a FoG-specific foundation model that detects FoG events and triggers interventions via Bluetooth. 
    }
    \label{fig:overview}
\end{figure}

The FM-FoG system aims to achieve real-time FoG detection through a novel foundation model-based approach that achieves generalization to unseen patients while maintaining energy efficiency for practical deployment. Our system integrates three key innovations: an energy-efficient event-triggered architecture, a context-aware foundation model specifically designed for IMU data, and FoG detection capabilities that eliminate patient-specific training requirements.

As shown in Figure \ref{fig:overview}, the system operates through a two-stage hierarchical pipeline designed to optimize both detection accuracy and energy consumption. Raw IMU data is continuously collected using UltiGesture (UG) devices \cite{Zhao2019Ultigesture} positioned on the ankle, providing non-intrusive placement that captures the most pronounced gait irregularities associated with FoG episodes. The quarter-sized form factor and wireless connectivity enable discreet, comfortable wear under regular clothing without impacting patient mobility or daily activities. User experience evaluation (conducted on all 23 participants) showed 97\% of participants reported positive experiences with device comfort, 3\% neutral, and 100\% expressed willingness to use the device for FoG management.

The first stage employs a lightweight CNN-LSTM activity trigger model that continuously monitors 1.28-second sliding windows at 100Hz to classify basic activity states. This trigger model serves as an energy-efficient gatekeeper, recognizing that FoG episodes occur exclusively during standing or walking activities. Since freezing cannot manifest during sedentary states such as sitting or lying, the system strategically activates the computationally intensive foundation model only when clinically relevant activities are detected.

Upon activation, the FoG-specific foundation model performs FoG episode detection through a context-aware transformer-based model that incorporates sensor placement information via embedding layers. The model utilizes knowledge gained during self-supervised pretraining on diverse movement datasets, then fine-tuned specifically for FoG detection through domain adaptation, enabling detection without requiring training on new patients' data.

When a FoG episode is detected, the system immediately triggers haptic intervention through vibration devices to provide cueing that helps patients overcome freezing episodes \cite{Klaver2023GoodVibrations, Sweeney2019WearableCueingDevices}. For our implementation, we use PDVibe3 \cite{Winfree2013StepSyncedVibration, Aggarwal2019StepSyncedVibrationTraining} vibrotactile intervention devices positioned at the ankle, delivering rhythmic pulsed vibrations in a 2-1 pattern (two seconds active vibration followed by one second rest) to minimize sensory adaptation while providing effective therapeutic cueing. The complete pipeline operates on a Google Pixel 8a smartphone, achieving intervention latency of approximately 20ms, which is well within the clinically acceptable timeframe for effective FoG intervention \cite{Lu2017EffectCueTiming}, ensuring timely therapeutic response while maintaining practical real-time performance.

In sum, the integrated system delivers a practical, energy-efficient solution for FoG management that addresses key limitations of existing approaches: eliminating patient-specific training requirements, reducing computational overhead through selective activation, and maintaining clinical effectiveness through real-time intervention. 

\begin{table*}[!htb]
  \centering
  \small
  \setlength{\tabcolsep}{4pt}
  \renewcommand{\arraystretch}{1.1}

  \caption{%
    Datasets used for pretraining the FM‐FoG model.
    This table summarizes the five wearable-sensor datasets employed to initialize our FoG-detection backbone. For each dataset we list sampling frequency, sensor placement, modalities, participant count, and label set.
  }
  \label{tab:datasets}

  \begin{tabularx}{\textwidth}{|p{0.12\textwidth}
                                |p{0.10\textwidth}
                                |p{0.15\textwidth}
                                |p{0.20\textwidth}
                                |p{0.13\textwidth}
                                |X|}
    \hline
    \textbf{Dataset}
      & \textbf{Frequency}
      & \textbf{Sensor Locations}
      & \textbf{Sensor Types}
      & \textbf{\# Subjects}
      & \textbf{Activities / Labels} \\
    \hline\hline

    tDCS FOG & 128Hz         & Lower back                   & Accelerometer                          
            & 67 PD patients           & FoG start hesitation, FoG turn, FoG walking, non-FoG         \\ \hline
    DeFOG   & 100Hz         & Lower back                   & Accelerometer                          
            & 66 PD patients           &  FoG, non-FoG                                        \\ \hline
    Daphnet & 64Hz          & Ankle, thigh, trunk          & Accelerometer                          
            & 10 PD patients         & FoG, non-FoG                                        \\ \hline
    PAMAP2  & 100Hz         & Wrist, chest, ankle          & \makecell[l]{Accelerometer, Gyroscope,\\Magnetometer}  
            & 9 healthy subjects            & 12 physical activities (lying, sitting, standing, walking, …)     \\ \hline
    Ours (VCU FoG-IMU)   & 100Hz         & Ankle                        & Accelerometer \& Gyroscope             
            & 23 PD patients           & FoG, non‐FoG                               \\ \hline
  \end{tabularx}
\end{table*}
\section{Methodology}
Our methodology encompasses four key components that enable effective FoG detection cross-patient generalization: comprehensive dataset collection and preprocessing, self-supervised pretraining with sensor context integration, fine-tuning for FoG-specific detection, and an event-triggered system design. This section details each component and shows how they collectively address the challenges of patient-specific training requirements and real-world deployment constraints.

\subsection{Datasets \& Preprocessing}
\subsubsection{Datasets}
Our FM-FoG system leverages a comprehensive collection of wearable sensor datasets to enable generalizable foundation model pretraining and effective FoG detection fine-tuning. As shown in Table \ref{tab:datasets}, we used five different datasets that collectively provide diverse movement patterns, sensor configurations, and patient populations necessary for learning generalizable representations.

We incorporated four established public datasets widely used in movement analysis research: (1) \textbf{tDCS FOG and DeFOG} \cite{tdcsfog, defog} provide PD-specific gait data from 133 patients with expert-annotated FoG episodes including start hesitation, turning, and walking-related freezing events in both clinical and daily-living settings; (2) \textbf{Daphnet} \cite{Roggen2010DaphnetFreezingOfGait} contains data from 10 PD patients with 237 annotated FoG episodes, providing relatively large gait variability across patients; (3) \textbf{PAMAP2} \cite{Reiss2012PAMAP2PhysicalActivity} extends our corpus beyond PD-specific data to include diverse physical activities from healthy subjects. This dataset contains recordings from 9 subjects performing 18 different physical activities including walking, running, cycling, and various daily activities. This dataset enhances the foundation model's understanding of general movement patterns and provides generalizable baseline representations for normal locomotion.




\vspace{1em}
\noindent\textbf{VCU FoG-IMU Dataset.} Our dataset consists of IMU data collected from 23 PD patients experiencing FoG episodes during daily activities. Data were collected using UltiGesture sensors positioned at the ankle, recording 3-axis accelerometer and gyroscope measurements at a 100Hz sampling frequency. Each participant was recruited with an established diagnosis of PD and confirmed FoG symptoms based on clinical assessment.

Data collection was conducted under institutional review board (IRB) approval, beginning in 2021 and ongoing. Table \ref{tab:dataset_summary} summarizes the demographic and recording characteristics of our dataset. Patients ranged in age from 57 to 82 years with participants of both genders. On average, each patient contributed 23 FoG episodes across $\approx 10$ minutes of walking data. In total, we recorded 528 FoG episodes over 237.9 minutes ($\approx 4$ hours) of annotated movement data.

\begin{table}[ht]
\centering
\caption{Summary of VCU FoG-IMU Dataset Characteristics. Including demographic statistics, FoG episode frequency, and recording duration across all 23 participants.}
\begin{tabular}{l l}
\toprule
\textbf{Characteristic} & \textbf{Value} \\
\midrule
Number of Patients        & 23 \\
Age Range (Mean ± SD)     &  69.9$\pm$5.7\\
Gender (M / F)            &  16/7\\
Avg. Height (cm$\pm$SD)   &  177.0$\pm$11.5\\
Avg. Weight (kg$\pm$SD)   &  98.5$\pm$6.1\\
Avg. FoG Episodes         &  23 per patient \\
Avg. Recording Duration   &  10.4 minutes \\
Total FoG Episodes        & 528 \\
Total Recording Duration  & 237.9 minutes ($\approx 4$ hours) \\
\bottomrule
\end{tabular}
\label{tab:dataset_summary}
\end{table}

To maximize freezing episode occurrence, clinical protocols incorporated five primary FoG-triggering conditions: dual-tasking, tight turns, narrow passage navigation, visual targets, and time-pressured walking \cite{Ishii2017CharacteristicsFoG}. Standard 10-meter straight-line walks provided baseline movement references. All sessions were conducted in controlled clinical environments with appropriate safety measures. FoG episodes were annotated by experienced movement disorder specialists through synchronized video analysis, providing ground-truth labels for both FoG onset and offset times. 

The patient cohort presents diverse motor symptoms including dyskinesia, dystonia, and ataxia, representing heterogeneous clinical presentations that make FoG detection challenging in real-world scenarios. Patients with concurrent dyskinesia exhibit rapid, involuntary movements that could potentially be misclassified as FoG, while those with dystonia present sustained muscle contractions affecting gait patterns. This symptom diversity challenges the model to distinguish FoG from other movement abnormalities, thereby improving its specificity for real-world deployment where patients commonly present with multiple concurrent motor symptoms.

For FoG detection, we segmented data into 1.28-second windows with 50\% overlap. This window size captures complete gait cycles even for slower walking patients while remaining short enough to avoid missing brief freezing episodes. For temporal labeling, we employed a conservative approach where windows containing freezing activity in the final 30\% of the time segment received positive labels, while all other windows were labeled as non-freezing. The resulting dataset contains 21,868 labeled windows, with approximately 30\% FoG episodes, enabling supervised fine-tuning for FoG detection tasks.

\vspace{1em}
\noindent\textbf{Combined Dataset for Foundation Model Pretraining.} For foundation model pretraining, we utilized all five datasets to maximize exposure to varied movement dynamics and sensor placement locations. This diverse pretraining corpus enables the model to learn transferable features that generalize across different sensor configurations and patient populations. Importantly, during the self-supervised pretraining phase, only the raw IMU sensor data are used without any FoG labels or annotations, ensuring that subsequent evaluation on our FoG dataset does not constitute data leakage. The pretraining process learns movement patterns through masked sequence reconstruction, making it fundamentally different from supervised learning that would require ground-truth labels.

The combined pretraining corpus exhibits several key characteristics relevant to IMU data: (1) \textbf{Diverse sampling frequencies} (64-128Hz), (2) \textbf{Varied sensor placements} across multiple body locations, necessitating spatial context awareness, (3) \textbf{Multiple sensor modalities} including accelerometer, gyroscope, and magnetometer data, and (4) \textbf{Subject diversity} spanning both FoG patients and healthy individuals across different age and gender groups, and varying disease severities.

\subsubsection{Preprocessing}
The heterogeneous nature of our multi-dataset corpus requires comprehensive preprocessing to enable effective training. Our preprocessing pipeline addresses four key challenges in our diverse IMU datasets.

\vspace{1em}
\noindent\textbf{Sampling Frequency Harmonization:} To address the diverse sampling rates across datasets, we implemented a unified resampling strategy targeting a consistent 100Hz sampling frequency. For upsampling lower-frequency data (such as Daphnet's 64Hz), we employed cubic spline interpolation \cite{kidger2020neuralcde} to generate intermediate data points while preserving the temporal smoothness of the original signal characteristics. This interpolation method maintains the continuity of acceleration and velocity profiles. For downsampling higher-frequency data (such as tDCS FOG's 128Hz), we applied random sub-sampling within desired time segments, which maintains the temporal representativeness of the original signal patterns.







\begin{table}[ht]
\centering
\caption{Impact of Resampling on FoG Detection Performance}
\resizebox{\linewidth}{!}{
\begin{tabular}{lcccc}
\toprule
\textbf{Method} & \textbf{F1 (\%) } & \textbf{Accuracy (\%)} & \textbf{Precision (\%)} & \textbf{Recall (\%)} \\
\midrule
With Resampling & 98.5 & 98.5 & 98.6 & 98.5 \\
No Resampling   & 91.2 & 91.4 & 91.9 & 91.4 \\
\bottomrule
\end{tabular}
}
\label{tab:resampling}
\end{table}

To quantify the effectiveness of our sampling frequency harmonization approach, we conducted ablation experiments comparing FoG detection performance with and without frequency resampling. Table \ref{tab:resampling} presents the results across multiple evaluation metrics.

The results show that frequency harmonization significantly improves FoG detection performance across all metrics. The resampling approach achieved a 7.3\% higher F1-score, a 7.1\% higher accuracy, and substantial improvements in precision and recall compared to using mixed-frequency data directly. This validates that standardizing sampling frequencies is essential for effective foundation model training, as heterogeneous temporal resolutions create inconsistent feature representations that hinder cross-dataset generalization. 

\vspace{1em}
\noindent\textbf{Sensor Placement Standardization:} Given the varied sensor placements across datasets (ankle, wrist, chest, thigh, trunk, and lower back), we developed a context-aware approach to handle spatial heterogeneity. Each sensor placement location is encoded through explicit location embeddings. Additionally, we standardized sensor axis directions across different devices and placements by applying appropriate coordinate transformations, ensuring that the X, Y, and Z axes maintain consistent orientations relative to physical directions regardless of the original device mounting configuration.

\vspace{1em}
\noindent\textbf{Multi-Modal Sensor Integration:} The datasets contain varying combinations of sensor modalities (accelerometer only, accelerometer plus gyroscope, or accelerometer plus gyroscope plus magnetometer). We established a standardized channel ordering protocol where accelerometer data (3 channels) is always positioned first, followed by gyroscope data (3 channels) when available, and magnetometer data (3 channels) when present. For datasets missing certain modalities, we created zero-padded channels to maintain consistent input dimensions across all data sources. Each modality undergoes modality-specific normalization to account for differing physical units and dynamic ranges, ensuring that no single sensor type dominates the learning process.

\vspace{1em}
\noindent\textbf{Comprehensive Data Normalization:} To mitigate individual variations in movement amplitude and sensor sensitivity while preserving clinically relevant patterns, we applied normalization at multiple levels. First, we conducted per-axis normalization to handle the varying distributions and scales across X, Y, and Z dimensions, as each axis captures different aspects of movement dynamics. Second, we applied per-subject normalization by computing subject-specific mean and standard deviation statistics across all recorded sessions for each individual, then applied z-score normalization to center each subject's data around zero mean with unit variance. Finally, we conducted foot-wise normalization for left and right ankle sensors independently, as gait patterns exhibit natural asymmetries and compensatory behaviors, particularly in PD patients where motor symptoms often manifest unilaterally or with different severity between sides. This asymmetry-aware normalization prevents the averaging out of clinically significant lateralized movement patterns that could be important for FoG detection.

The complete preprocessing pipeline transforms the heterogeneous multi-dataset corpus into a unified, standardized format suitable for foundation model training while preserving the clinically relevant movement patterns in the IMU data.

\subsection{Self-supervised Pretraining with Sensor Context Integration}
The proposed FM-FoG pretraining uses a novel self-supervised learning approach that combines masked sequence reconstruction with explicit sensor context integration, enabling the model to learn, generalizable representations from unlabeled IMU data with encoded spatial and configurational properties of different sensor placements. This addresses a critical challenge in IMU data generalization where sensors may be positioned at varying body locations with different orientations.

\begin{figure}[!t]
    \centering
    \includegraphics[width=\linewidth, trim=17cm 3cm 17cm 3.5cm, clip]{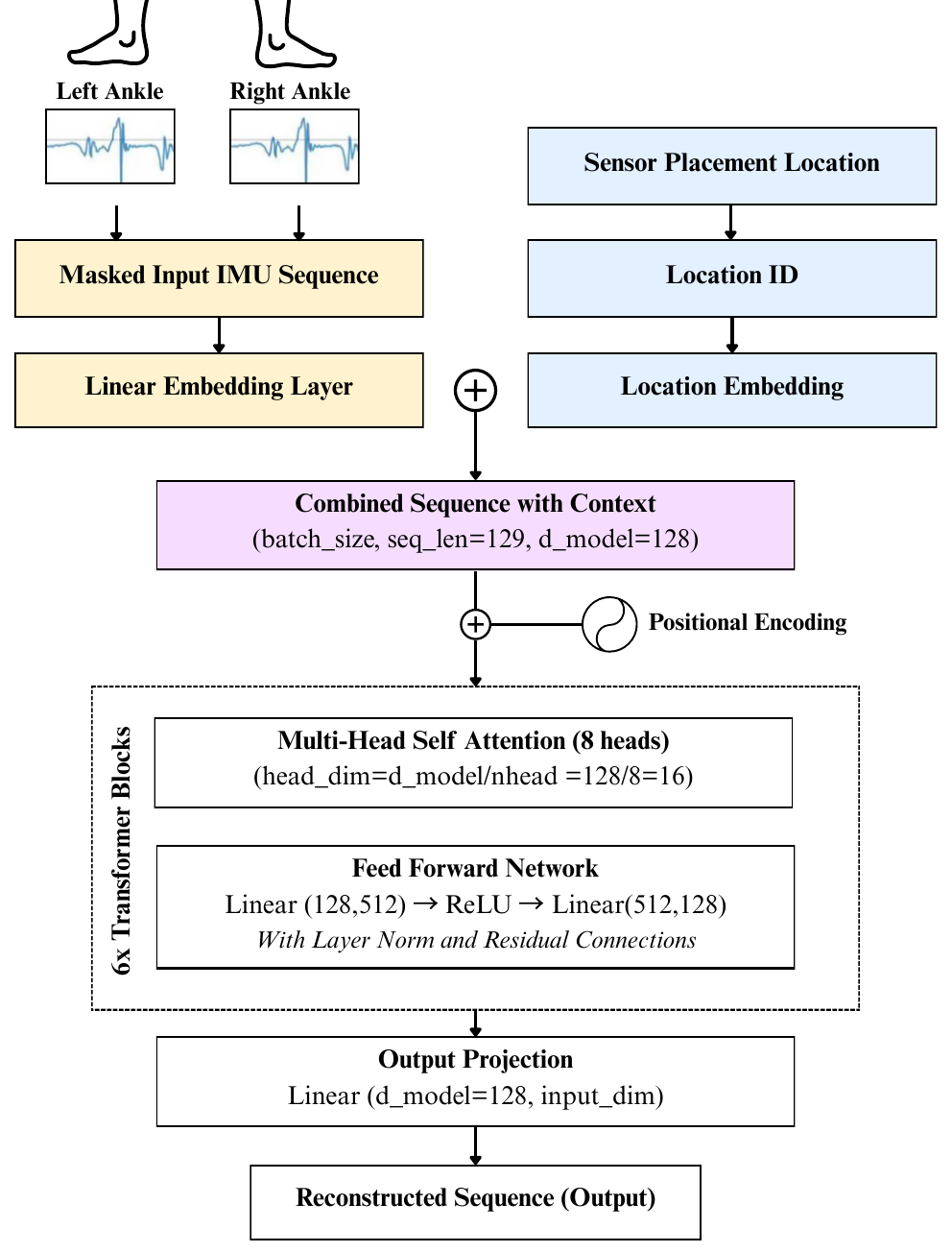}
    \caption{Pretraining pipeline for masked IMU reconstruction with a 6-block Transformer. Masked IMU sequences are combined with learnable sensor placement context information, then processed through the pipeline, with the final output reconstructing the original IMU values.}
    \label{fig:pretrain}
\end{figure}

\textbf{Overview:} The proposed FM-FoG uses an encoder-based transformer \cite{Vaswani2017Attention} specifically designed for IMU time series data with integrated sensor context awareness. As illustrated in Fig. \ref{fig:pretrain}, the model processes raw IMU sequences through a dual-pathway approach: the primary pathway handles the temporal sensor data through linear embedding, while the auxiliary pathway processes sensor placement information through learned location embeddings. Positional encoding is applied to preserve temporal ordering, and the combined sequences are processed through transformer blocks (8 attention heads, model dimension 128). Each transformer block incorporates layer normalization, residual connections, and feed-forward networks with ReLU activation.

\textbf{Self-supervised Learning with Masked Reconstruction: }We use a masked sequence reconstruction pretext task \cite{devlin2019bertpretrainingdeepbidirectional} that encourages the model to learn meaningful temporal dependencies in IMU data. Input sequences of 1.28 seconds (128 data points at 100Hz) are processed with 30\% of positions randomly masked. The linear embedding layer projects multi-modal sensor data to the model dimension, and after transformer processing, an output projection layer maps representations back to the original sensor space. The reconstruction loss is computed using mean squared error between predicted and actual sensor values at masked positions, providing a strong learning signal for temporal pattern modeling.

\textbf{Sensor Context Integration:} A key innovation of our approach is the explicit modeling of sensor placement context through learnable embeddings. Each sensor placement location (ankle, thigh, lower back, etc.) is mapped to a unique location ID, which is transformed into a dense vector representation through a learned embedding layer. This location embedding captures the spatial and biomechanical properties associated with each sensor position, allowing the model to understand how movement patterns should be interpreted differently based on sensor placement. The location embedding is then combined with the linearly embedded IMU sequence, providing contextual information that influences all subsequent transformer computations and facilitates effective transfer to new sensor configurations during fine-tuning.

The model was trained using the AdamW optimizer \cite{Loshchilov2017DecoupledWeightDecayRegularization} with a learning rate of $5\times10^{-4}$, weight decay of 0.01, and a cosine annealing scheduler. Training used mixed precision with batch size of 32 for 50 epochs across the combined dataset corpus.

This self-supervised pretraining approach establishes a strong foundation for subsequent fine-tuning on FoG detection tasks, as it provides the model with rich, generalizable representations of human movement patterns while maintaining sensitivity to spatial and configurational factors that influence sensor measurements across different deployment contexts.

\subsection{Fine-tuning FM-FoG}
Following self-supervised pretraining, we then fine-tune the proposed {FM-FoG} on a small subset of labeled clinical data. The pretrained transformer model serves as the backbone, on which we perform full parameter fine-tuning. We add a classification head consisting of global average pooling, a linear classifier, and softmax activation to produce binary FoG/non-FoG predictions.

The fine-tuning uses different learning rates to balance knowledge retention and task adaptation: transformer layers use a reduced learning rate of $1\times10^{-4}$ to preserve learned movement representations, while the classification head uses $1\times10^{-3}$ for rapid convergence. Data augmentation through temporal jittering and Gaussian noise injection are used to improve robustness.

The fine-tuning process addresses the domain shift from general movement patterns to PD-specific gait abnormalities, leveraging the sensor context integration capabilities built during pretraining. Our approach allows the model to develop specialized sensitivity for accurate real-time FoG detection while preserving its comprehensive movement understanding.

\subsection{Event-Triggered FM-FoG}
We propose an event-triggered architecture using a lightweight activity classifier as a gatekeeper, significantly reducing energy consumption.

\begin{figure}[!t]
    \centering
    \includegraphics[width=\linewidth, trim=4.8cm 7.8cm 5.95cm 7.75cm, clip]{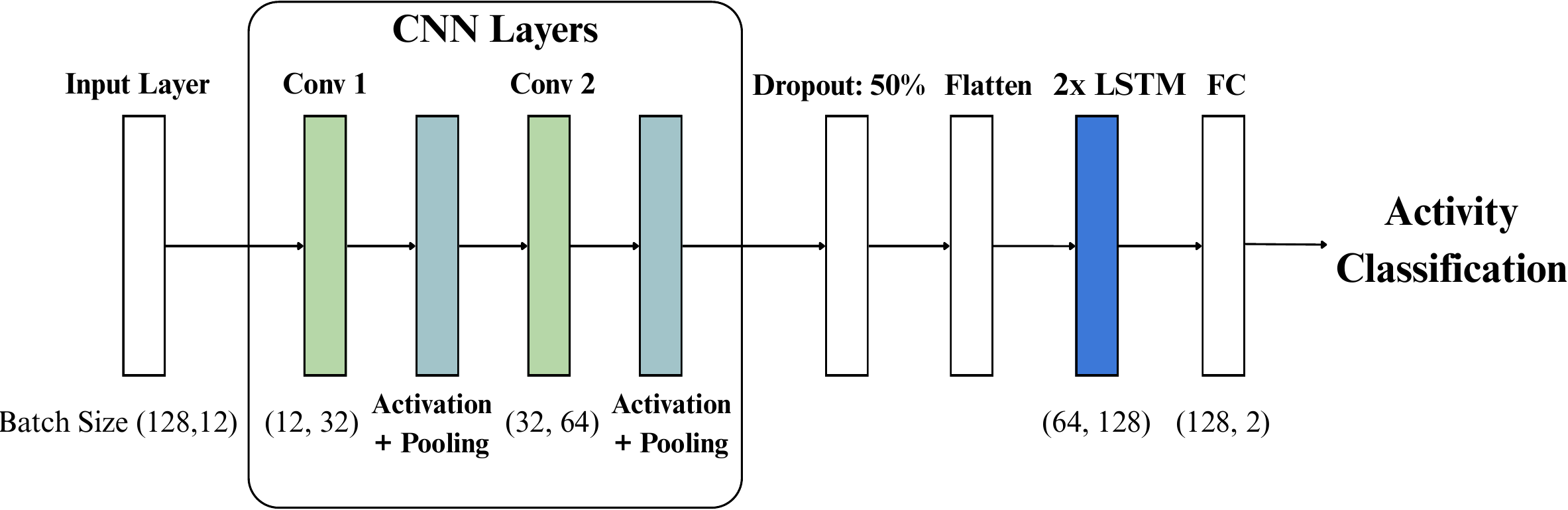}
    \caption{The event-triggering activity classifier: a two-stage CNN-LSTM model.}
    \label{fig:cnn}
\end{figure}

The activity trigger classifier uses a small CNN-Long Short-Term Memory (LSTM) model with $\approx 5$MB in memory 
specifically optimized for efficient activity classification in PD patients, as shown in Fig. \ref{fig:cnn}. The model processes 1.28-second IMU windows through two convolutional layers (32 and 64 filters with kernel size 12) with ReLU activation and max pooling, followed by 50\% dropout for regularization. The extracted features are then processed by a two-layer LSTM (64 hidden units each) to capture temporal dependencies, followed by a fully connected layer that outputs binary classification probabilities for standing/walking versus sitting/lying activities.

The trigger model continuously monitors raw IMU data and classifies user states into two categories: FoG-relevant states (standing/walking, including attempts at walking where users appear to be stationary but are actually trying to move) and FoG-irrelevant states (sitting/lying). This classification recognizes that FoG episodes occur only during ambulatory activities -- patients cannot experience FoG while seated or lying down, as these postures do not involve the gait mechanisms that can freeze. FoG specifically affects gait movements while supporting body weight. The proposed FM-FoG is activated only when FoG-relevant states are detected, as freezing episodes cannot manifest during non-ambulatory activities.


\section{Evaluation}
We evaluate FM-FoG's performance through comprehensive experiments designed to demonstrate its cross-patient generalization capabilities and practical deployment advantages.
\subsection{Experimental Settings}
To establish the cross-patient generalization capabilities of our approach, we partitioned the VCU FoG-IMU of 23 PD patients into two distinct groups: 16 patients (randomly selected) for fine-tuning the proposed FM-FoG,
and the remaining 7 patients as a completely unseen test set. This experimental design ensures that the performance of the model in test subjects represents a true generalization between patients without requiring individual calibration. We repeated this random partitioning 25 times to obtain statistical evaluation and to account for potential variance in patient selection.

\subsubsection{Baseline Methods} 
We compare FM-FoG against three categories of state-of-the-art approaches. First, we evaluate against domain-specific FoG detection models, including Gait-Guard \cite{Gait-Guard} and the recent Dual-Level FoG recognition model \cite{Chen2025SoTA}. Second, we assess performance against general-purpose large language models, including GPT-4o \cite{OpenAI2024GPT4o} and LLaMA2-7b \cite{touvron2023llama2} to demonstrate the necessity of domain-specific foundation models for IMU data. Third, we compare our approach with existing time-series foundation models, including MOMENT \cite{MOMENT} and TimesFM \cite{TimesFM} to validate our specialized approach for wearable sensor healthcare applications. 

\subsubsection{Environmental Settings} 
All models were implemented using a standardized training pipeline to ensure fair comparison, with training performed on an NVIDIA A6000 GPU and hyperparameters tuned to maximize performance. Due to computational constraints and model size limitations, only the FoG-specific models were deployed and evaluated on the mobile device, while larger foundation models were assessed in controlled computational environments. All deployable models were evaluated on a Google Pixel 8a smartphone to assess real-world performance under practical hardware constraints. This consumer-grade device provides a realistic testbed for evaluating the energy efficiency and computational feasibility of different approaches in actual deployment scenarios.

\subsubsection{Evaluation Metrics} We assessed model performance using standard classification metrics including F1-score, accuracy, precision, and recall to evaluate FoG detection quality. Additionally, we measured practical system deployment considerations including model size, inference time, and resource consumption (energy, computation, and storage) to demonstrate real-world viability.

\begin{table*}[ht]
\centering
\caption{Comparative Performance and Parameter Counts}
\begin{tabular}{lccccr}
\toprule
Model                                 & F1\,($\mu\pm\sigma$) (\%)  & Accuracy\,($\mu\pm\sigma$) (\%) & Precision\,($\mu\pm\sigma$) (\%) & Recall\,($\mu\pm\sigma$) (\%) & Model Size                                                    \\
\midrule
Gait-Guard \cite{Gait-Guard} & 90.9$\pm$1.3 & 91.3$\pm$1.1 & 91.2$\pm$1.0 & 91.3$\pm$1.1 & 1.5M \\
Dual-Level \cite{Chen2025SoTA} & 71.9$\pm$7.5 & 72.1$\pm$7.7 & 73.8$\pm$7.8 & 70.8$\pm$7.9 & \textbf{0.17M} \\
GPT-4o \cite{OpenAI2024GPT4o} & 36.1$\pm$6.4 & 39.0$\pm$1.8 & 52.7$\pm$7.7 & 39.0$\pm$1.8 & 1.76T \\
LLaMA2-7b \cite{touvron2023llama2} & 59.1$\pm$0.0 & 71.1$\pm$0.0 & 50.5$\pm$0.0 & 71.1$\pm$0.0 & 7B \\
MOMENT-base \cite{MOMENT} & 85.9$\pm$3.8 & 86.3$\pm$3.3 & 86.8$\pm$3.2 & 86.3$\pm$3.3 & 113M                     \\
TimesFM-200M \cite{TimesFM, partai2025flamingotimesfm} & 83.9$\pm$8.3 & 89.9$\pm$2.2 & 78.9$\pm$ 10.1 & 89.7$\pm$4.9 & 200M \\
TimesFM-500M \cite{TimesFM, partai2025flamingotimesfm} & 84.1$\pm$13.8 & 93.9$\pm$2.0 & 77.9$\pm$15.9 & 92.0$\pm$10.0 & 500M \\
\textbf{Ours} & \textbf{98.5$\pm$0.7} & \textbf{98.5$\pm$0.7} & \textbf{98.6$\pm$0.7} & \textbf{98.5$\pm$0.7} & 1.2M \\
\bottomrule
\end{tabular}
\label{tab:performance}
\end{table*}

\subsection{Cross-Patient Detection Performance}
We evaluated FM-FoG's FoG detection capabilities on unseen patients against three categories of state-of-the-art approaches: FoG-specific detection models, general-purpose LLMs, and time-series FMs. We treat each individual patient as a separate domain. All evaluations for FoG-specific detection models and time-series FMs followed the same experimental protocol with 16 randomly selected patients from VCU FoG-IMU for training and 7 unseen patients for testing, repeated across 25 random train/test splits to ensure statistical analysis. For GPT-4o, API-based evaluations were conducted five times using identical inputs to ensure fair comparisons. For LLaMA2, we used the pretrained checkpoint for inference without any fine-tuning, which explains the absence of standard deviation in our results, as the model parameters remain fixed and produce identical outputs for our classification task.

For fair and accurate comparisons, we contacted the authors of Gait-Guard and obtained their exact implementation, enabling precise replication of their approach under our experimental conditions. For the Dual-Level model, we carefully implemented their methodology following the detailed algorithmic descriptions provided in their publication. FoG-specific models and time-series FMs were fine-tuned using identical protocols as our approach, while LLMs were evaluated in their reference configuration due to computational constraints preventing fine-tuning of billion-parameter models.

As shown in Table \ref{tab:performance}, FM-FoG achieves superior performance across all metrics, with 98.5\% F1-score, 98.5\% accuracy, 98.6\% precision, and 98.5\% recall, outperforming all baseline approaches. The results reveal a clear performance hierarchy: FoG-specific detection models (FM-FoG: 98.5\% F1, Gait-Guard: 90.9\% F1) significantly outperform time-series FMs (MOMENT: 85.9\% F1, TimesFM-500M: 84.1\% F1), which in turn exceed general-purpose LLMs (LLaMA2-7b: 59.05\% F1, GPT-4o: 36.1\% F1). This shows the importance of domain-specific model design for high-reliability healthcare applications. 

While both Gait-Guard and Dual-Level represent recent advances in FoG detection, their performance differs significantly. Gait-Guard achieves competitive results with 90.9\% F1, demonstrating the effectiveness of turn-aware feature engineering and multi-input transformer models for FoG detection. However, the Dual-Level model's performance is limited by its training methodology, achieving 71.9\% F1. This approach treats any non-normal walking gait patterns of PD patients as FoG, without accounting for other neurological conditions that cause gait abnormalities, such as ataxia, dystonia, twisted ankles, or general balance dysfunction. This limitation becomes particularly critical when evaluating on ankle-mounted IMU data, as ankle placement is highly sensitive to various gait-related disorders beyond FoG. Our patient cohort exhibits diverse motor symptoms, making binary normal/abnormal classification insufficient for real-world clinical scenarios.

Despite having substantially more parameters, time-series FMs underperformed compared to our specialized approach. MOMENT was pretrained on diverse time-series datasets including vehicle motion and gesture recognition data, which, while motion-related, fundamentally differ from IMU-based gait analysis in terms of sensor characteristics, movement patterns, and clinical relevance. Similarly, TimesFM's pretraining on web data such as Google search trends and Wikipedia page views, though extensive, lacks relevance to physiological movement patterns and healthcare applications. In contrast, FM-FoG's focused pretraining exclusively on IMU gait data from multiple sensor placements enables superior learning for FoG detection. The substantial performance gap between FM-FoG and time-series FMs highlights the importance of domain-specific pretraining for IMU data. While MOMENT and TimesFM leverage massive parameter counts, their exposure to data domains irrelevant to movements and healthcare dilutes their effectiveness for physiological signal analysis. This confirms that targeted, high-quality domain-specific pretraining outperforms general-purpose large-scale pretraining for specialized healthcare tasks, even with fewer total parameters.

Beyond performance superiority, FM-FoG maintains practical deployment advantages with a compact 1.2M parameter model compared to the substantially larger LLMs and FMs. This efficiency makes real-time mobile deployment feasible for practical use.

\subsection{System Performance}
We evaluated the real-world deployment performance of FM-FoG on a Google Pixel 8a smartphone \cite{Google2024Pixel8a} via Android Studio \cite{androidstudio} using the VCU FoG-IMU dataset. The Google Pixel 8a is a budget-oriented device with known battery limitations \cite{androidpolice_pixel8a_2024, linustech_pixel8a_2024, trustedreviews_pixel8a_2024}, which should be considered when interpreting our battery life results.

The Android implementation uses Bluetooth Low Energy (BLE) \cite{BluetoothSIG2016BLE} for sensor communication with automatic reconnection handling. Our analysis encompasses energy consumption, computational resource utilization, and system latency to show the practical viability of our proposed event-triggered FM-FoG. We primarily compare against Gait-Guard as it is the only FoG detection system with demonstrated mobile deployment in prior literature. LLMs and time-series FMs exceed mobile hardware constraints due to their large parameter counts, while other recent approaches such as the Dual-Level model have not demonstrated mobile deployment capabilities in their publications.

\begin{figure*}[!htbp]
  \centering
\includegraphics[width=0.8\linewidth, trim=0cm 0cm 0cm 0.85cm, clip]{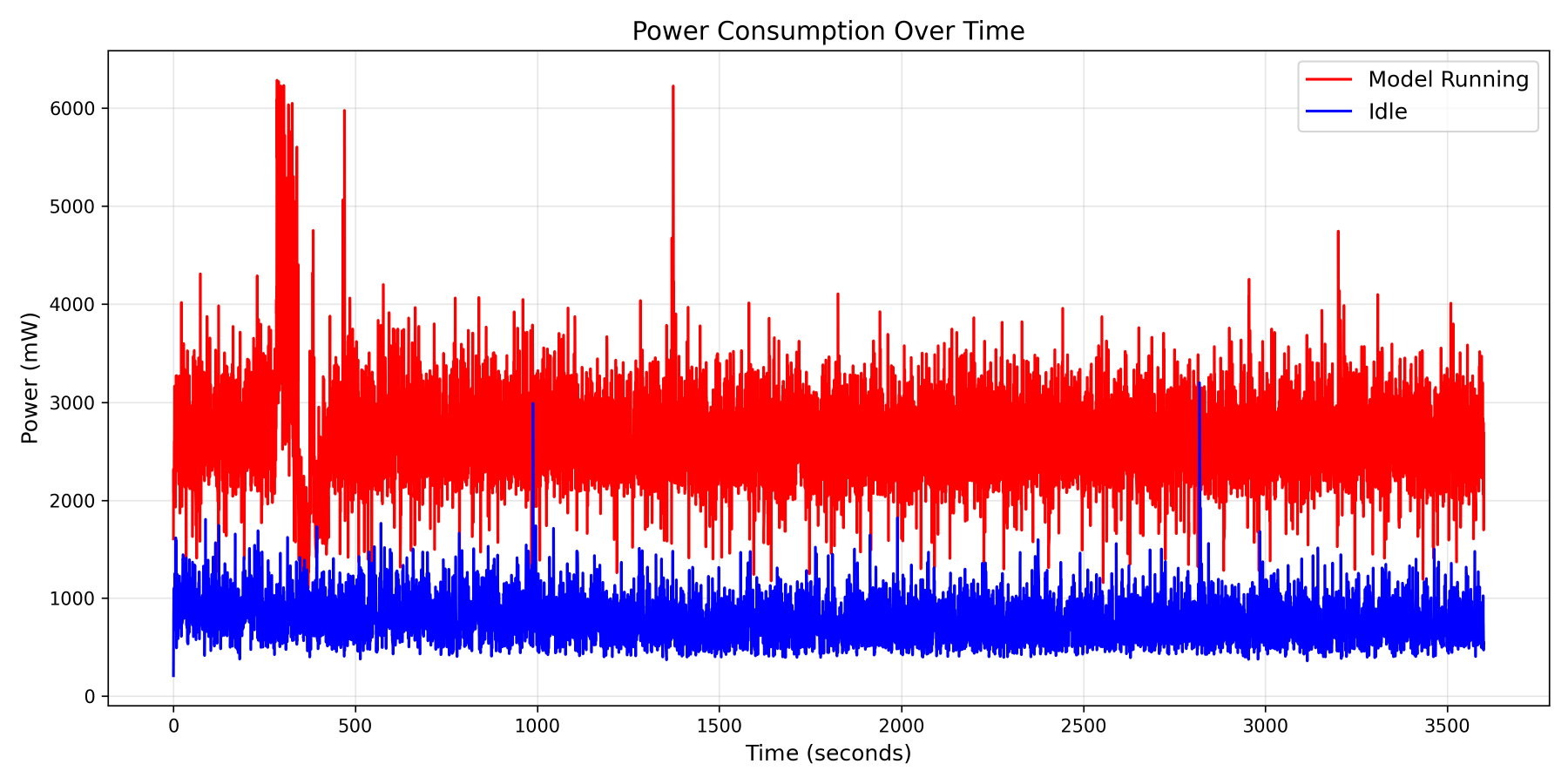}
  \caption{Real-time power consumption trace during FM-FoG model running continuously v.s. idle smartphone operation.}
  \label{fig:power_trace}
\end{figure*}

\begin{table}[!htbp]
  \centering
  \setlength{\tabcolsep}{13pt}
  \caption{Real-time performance comparison on Google Pixel 8a. Inference time measures model processing duration, while intervention time includes complete end-to-end response from data processing to vibration intervention activation.}
  \begin{tabular}{@{}l c c@{}}
    \toprule
    Model & \makecell{Inference Time \\(ms) $\downarrow$} & \makecell{Intervention Time \\(ms) $\downarrow$} \\ 
    \midrule
    Gait-Guard & 15.7 & 18.9\\
    FM-FoG & 14.3 & 17.5\\
    Triggered FM-FoG & 16.7 & 19.9 \\
    \bottomrule
  \end{tabular}
  \label{tab:inference-times}
\end{table}

\subsubsection{Latency} As shown in Table \ref{tab:inference-times}, FM-FoG shows competitive inference performance compared to the existing Gait-Guard approach. The standalone FoG detection foundation model (FM-FoG) achieves a 14.3ms inference time with a 17.5ms total intervention latency, slightly outperforming Gait-Guard's 15.7ms and 18.9ms respectively. When deployed with the CNN-LSTM activity triggering model, the complete system maintains real-time performance with a 16.7ms inference time and a 19.9ms intervention latency. These sub-20ms response times ensure timely intervention delivery. 

\subsubsection{Energy Consumption} 

\textbf{Power Measurement Methodology.} Accurate power measurement on mobile devices presents unique challenges, as USB-connected power monitoring can interfere with natural battery discharge patterns. To address this, we used a wireless power monitoring approach using Android Debug Bridge (ADB) over WiFi connection. However, our FoG detection system operates independently using only Bluetooth communication and does not require WiFi connectivity for normal model running. All current power measurements include WiFi constantly transmitting data for measurement purpose, which consumes significant power. This WiFi overhead is not necessary for our actual FoG detection model, so the true power consumption of our system would be substantially lower than reported values.

\begin{figure}[!htbp]
\centering
\includegraphics[width=\linewidth, trim=0.25cm 0.25cm 0cm 0.75cm, clip]{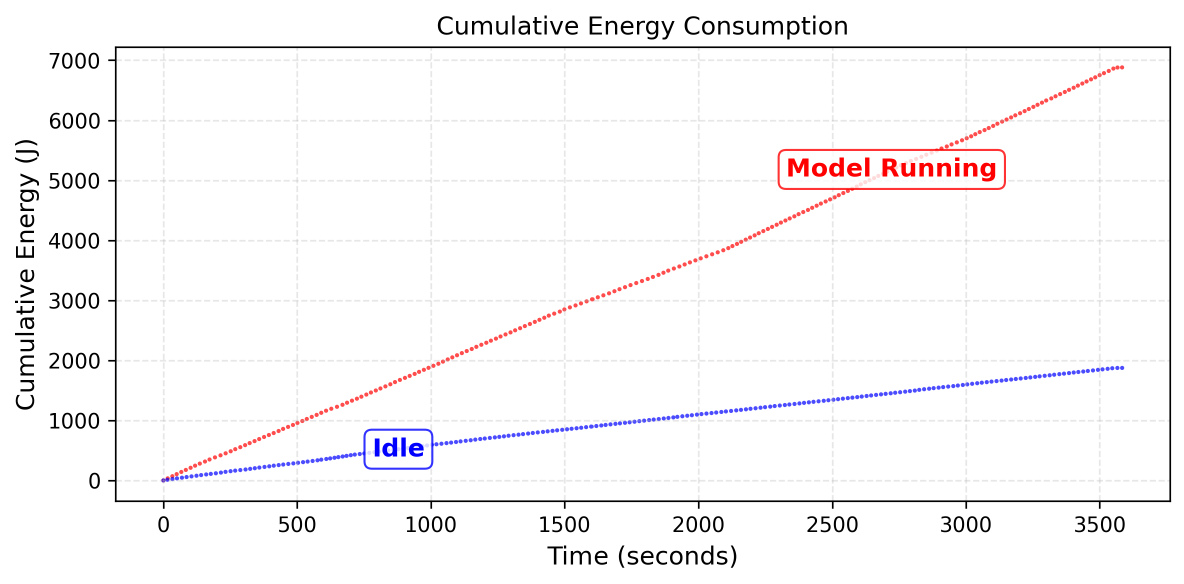}
\caption{Cumulative energy consumption over one-hour operation. The linear relationship shows measurement accuracy and system stability. Slopes validate power measurements.}
\label{fig:cumulative}
\end{figure}

Our power monitoring method samples battery current and voltage at 100ms intervals using Android's power supply interface, accessed through the Linux file-system at \textit{/sys/class/power\_supply\\/battery/}. The measurement process involves: (1) resetting battery statistics to establish discharge mode, (2) continuous sampling of instantaneous current ($\mu A$) and voltage ($\mu V$), (3) calculating instantaneous power as $P = |I|\times V$, and (4) integrating power over time to determine energy consumption.

\vspace{1em}
\begin{table}[!htbp]
  \centering
  \setlength{\tabcolsep}{7pt}
  \caption{Power consumption and battery life. Triggered rates represent the percentage of time the FM-FoG model is active.}
  \begin{tabular}{@{}l c c c c@{}}
    \toprule
    State & Power & Battery Life\\
    \midrule
    Idle & 724.2mW (0.7W) & 24.7h \\
    Gait-Guard & 2.8W & 6.2h \\
    FM-FoG & 2.6W & 6.7h \\
    \makecell[l]{Triggered FM-FoG \\(10\% triggered)} & 1.5W & 11.5h \\
    \makecell[l]{Triggered FM-FoG \\(20\% triggered)} & 1.7W & 10.2h \\
    \makecell[l]{Triggered FM-FoG \\(30\% triggered)} & 1.8W & 9.6h \\
    \makecell[l]{Triggered FM-FoG \\(40\% triggered)} & 2.1W & 8.2h \\
    \makecell[l]{Triggered FM-FoG \\(50\% triggered)} & 2.3W & 7.5h \\
    \makecell[l]{Triggered FM-FoG \\(60\% triggered)} & 2.4W & 7.2h \\
    \makecell[l]{Triggered FM-FoG \\(70\% triggered)} & 2.7W & 6.4h \\
    \makecell[l]{Triggered FM-FoG \\(80\% triggered)} & 2.9W & 6h \\
    \makecell[l]{Triggered FM-FoG \\(90\% triggered)} & 2.9W & 6h \\
    \bottomrule
  \end{tabular}
  \label{tab:energy}
\end{table}
\begin{figure}[!t]
\centering
\includegraphics[width=0.8\linewidth, trim=0.25cm 0.25cm 0.25cm 1.25cm, clip]{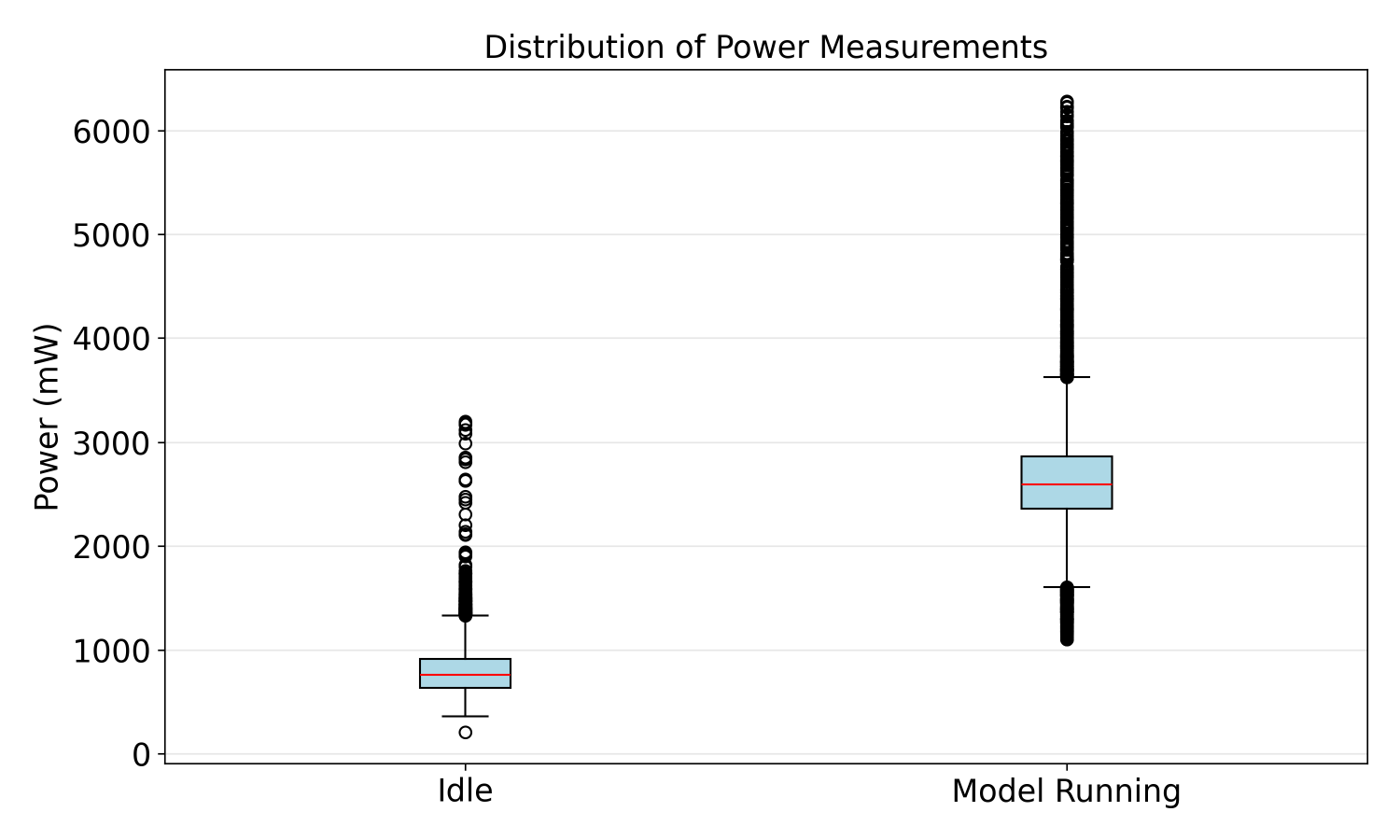}
\caption{Power consumption distribution during different states. Box plots show median, quartiles, and outliers across 3600 one-second measurements.}
\label{fig:boxplot}
\end{figure}
\noindent\textbf{Real-time Power Consumption.} To validate our measurement methodology and show system behavior, we conducted controlled experiments comparing FM-FoG power consumption against idle smartphone operation.

Fig. \ref{fig:power_trace} presents real-time power consumption measurements comparing two operational states: the FM-FoG foundation model running continuous inference versus the smartphone in idle state with no background apps running, screen off, and WiFi and Bluetooth enabled. The real-time power consumption comparison shows FM-FoG consuming a steady $\approx 2.6$W while idle operation consumes $\approx 0.7$W. The linear relationship in cumulative energy consumption (Fig. \ref{fig:cumulative}) confirms measurement accuracy and system stability, with slopes directly corresponding to average power consumption under both states.
\begin{table}[!t]
  \centering
  \setlength{\tabcolsep}{7pt}
  \caption{System resource consumption. CPU usage is reported as percentage of total available cores (9 cores).}
  \begin{tabular}{@{}l c c c c@{}}
    \toprule
    State & CPU & Memory \\
    \midrule
    Gait-Guard & 18.1\% & 112MB \\
    FM-FoG & 16.1\% & 107MB \\
    \makecell[l]{Triggered FM-FoG \\(10\% triggered)} & 16.1\% & 112MB \\
    \makecell[l]{Triggered FM-FoG \\(20\% triggered)} & 16.1\% & 112MB \\
    \makecell[l]{Triggered FM-FoG \\(30\% triggered)} & 16.2\% & 112MB \\
    \makecell[l]{Triggered FM-FoG \\(40\% triggered)} & 16.3\% & 112MB \\
    \makecell[l]{Triggered FM-FoG \\(50\% triggered)} & 16.4\% & 112MB \\
    \makecell[l]{Triggered FM-FoG \\(60\% triggered)} & 16.5\% & 112MB \\
    \makecell[l]{Triggered FM-FoG \\(70\% triggered)} & 16.6\% & 112MB \\
    \makecell[l]{Triggered FM-FoG \\(80\% triggered)} & 16.7\% & 112MB \\
    \makecell[l]{Triggered FM-FoG \\(90\% triggered)} & 17.1\% & 112MB \\
    \bottomrule
  \end{tabular}
  \label{tab:resource-usage}
\end{table}
\begin{figure*}[!h]
    \centering
    \includegraphics[width=0.8\linewidth, trim=7cm 2cm 7cm 2cm, clip]{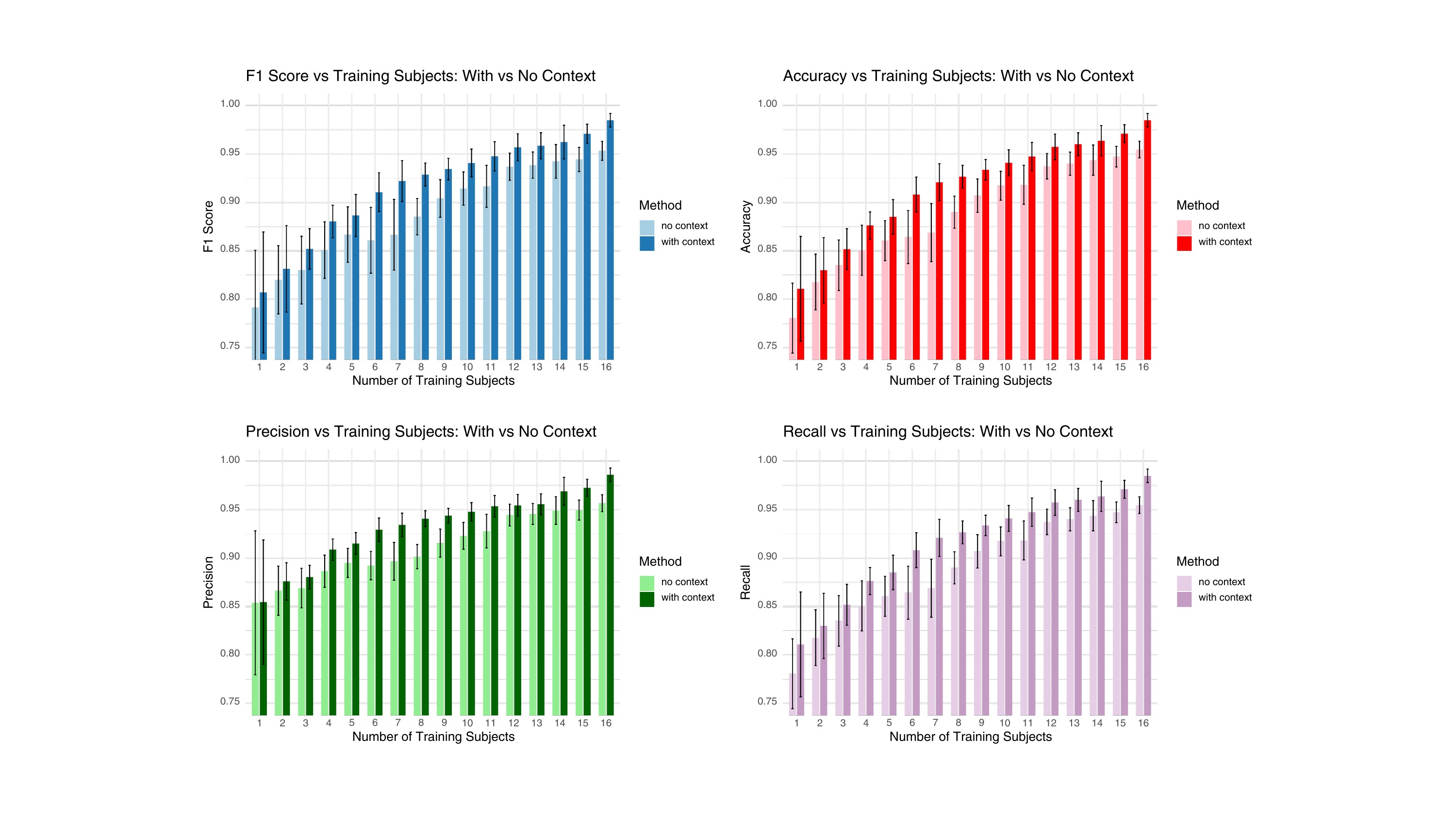}
    \caption{Ablation study results comparing FoG detection performance on unseen patients with and without sensor context integration across varying training set sizes. Results show F1-score, accuracy, precision, and recall when training on 1-16 patients from VCU FoG-IMU and testing on remaining unseen subjects. Models show consistent improvements and reduced variance, with context-aware models (darker bars) consistently outperform baseline models without context information (lighter bars).}
    \label{fig:ablation}
\end{figure*}
Fig. \ref{fig:boxplot} shows the statistical distribution of power measurements across 3600 seconds (1 hour), revealing FM-FoG's consistent computational load with reasonable variance suitable for daily-use applications. The box plots show that our FM-FoG maintains predictable energy demands without excessive power spikes that could impact user experience or device thermal management. 

\vspace{1em}

\noindent\textbf{Energy Efficiency Results.} 
The event-triggered system delivers substantial energy savings compared to continuous monitoring approaches, as shown in Table \ref{tab:energy}. Clinical evidence indicates that PD patients experiencing FoG are typically 60+ years old \cite{parkinsonsfoundation_stats, hopkinsmedicine_parkinsons_dementia}. Research in elderly populations shows that adults 65 and older spend approximately 65\% of their waking time sedentary (8.5-9.6 hours daily) \cite{wullems2016sedentarism}. This translates to approximately 35\% of waking time spent in ambulatory states (standing/walking), which corresponds to our 30-40\% triggering rate scenarios in Table \ref{tab:energy}. Since FoG occurs exclusively during walking or standing activities, our triggering strategy aligns directly with real-world activity patterns.

The event-triggered system delivers significant energy savings compared to continuous monitoring approaches. At the realistic 30\% triggering rate, the event-triggered FM-FoG achieves 9.6 hours of battery life compared to 6.7 hours for continuous detection, representing a 43\% improvement. Even conservative estimates at 40\% triggering rate maintain 8.2 hours of monitoring, while minimal activity scenarios (10\% triggering) extend battery life to 11.5 hours, substantially improving practical deployment duration across diverse patient activity levels.

\subsubsection{Computational Resource Utilization}
System monitoring reveals efficient resource management across all deployment scenarios in Table \ref{tab:resource-usage}. Both the CNN-LSTM activity trigger model and the FM-FoG model are permanently loaded in memory to enable instantaneous switching without model loading overhead, which would introduce excessive latency for real-time intervention systems. Dynamic model loading/unloading would require file I/O operations, memory reallocation, and model initialization that could delay critical FoG interventions. This persistent loading strategy explains the consistent 112MB memory footprint across all triggering rates, as memory allocation remains constant. The consistent 112MB memory footprint ensures compatibility with mid-range smartphones, making the system accessible to a broader patient population.

CPU usage remains stable at 16.1-17.1\% across different triggering rates, slightly outperforming Gait-Guard's 18.1\% usage. The slight CPU variation reflects the dynamic workload: the lightweight CNN-LSTM model runs continuous inference for activity classification, while the computationally intensive foundation model activates only when standing/walking is detected. The modest increases in memory footprint and CPU utilization are justified by the significant battery life improvements, representing an optimal trade-off between computational resources and energy efficiency. The stable CPU usage profile indicates sustainable long-term operation without thermal throttling concerns.

\subsection{Ablation Studies 
}
We conducted ablation studies to evaluate the impact of sensor context integration on cross-patient FoG detection performance. The ablation focused on comparing our proposed context-aware FM-FoG against the model without sensor placement information.

We systematically varied the number of training subjects from 1 to 16 (1 patient's data to $\frac{2}{3}$ of our patients' data) and evaluated performance on the remaining unseen subjects. For each training size, we compared two model variants: our complete approach with sensor context embeddings (``with context'') and a baseline version without spatial context information (``no context''). This design allowed us to isolate the contribution of context awareness.

As shown in Fig. \ref{fig:ablation}, the results show substantial improvements when incorporating sensor context information across all metrics and training set sizes. With context integration, the model achieves higher F1-scores (0.81-0.99 vs. 0.76-0.96), accuracy (0.83-0.99 vs. 0.81-0.99), precision (0.85-0.99 vs. 0.82-0.99), and recall (0.84-0.99 vs. 0.82-0.99) compared to the no-context baseline. Notably, the context-aware approach shows more pronounced improvements in smaller training regimes, with F1-score differences of approximately 5-8\% when using fewer than 8 training subjects.

The context-aware model shows remarkable cross-patient learning capabilities, achieving $> 90\%$ F1-score with approximately 6-7 training subjects (approximately $\frac{1}{3}$ of the patients), representing a critical inflection point where performance becomes highly reliable. Furthermore, with around 12 training subjects, the model consistently exceeds 95\% F1-score, demonstrating that effective FoG detection that generalizes to new patients can be achieved with data from only half of the patient population. This efficiency in data utilization shows FM-FoG's ability to extract generalizable movement patterns that transfer effectively across unseen patients. 

Beyond improved absolute performance, sensor context integration significantly reduces performance variance across different patient selections. The error bars show smaller confidence intervals for the context-aware model, indicating more stable and reliable generalization regardless of the specific training subjects selected.
\section{Discussion \& Future Work}
\subsection{Implications of Domain-Specific Foundation Models for Healthcare AI}
Our results show that large, general-purpose models do not necessarily outperform specialized alternatives in healthcare applications. The superior performance of our model with 1.2M parameters against larger time-series FMs suggests that thoughtful domain-specific design can be more effective than scale alone. This has broader implications for healthcare AI development, where resource constraints and interpretability requirements often favor smaller, more targeted approaches. The principle extends beyond movement disorders to other physiological monitoring applications where signal characteristics and clinical context differ fundamentally from general-purpose AI domains. The cross-patient generalization capability of our model addresses the challenge where new patients cannot benefit from existing AI systems until sufficient individual data is collected. The ability to provide immediate FoG detection without patient-specific training could significantly reduce healthcare costs and improve accessibility for underserved populations. This approach enables rapid deployment in clinical settings where traditional systems would require weeks of data collection and calibration, during which patients remain vulnerable to falls and mobility decline.

\subsection{Longitudinal Monitoring}
Our dataset primarily focuses on controlled clinical sessions rather than extended daily monitoring. The lack of diverse daily activity data limits understanding of system performance across the full spectrum of patient behaviors and temporal variations in PD symptoms.

Longitudinal monitoring studies using multiple wearable sensors have shown feasibility for large-scale deployment in Parkinson's disease, with research showing that continuous, objective data collection in real-world settings is essential for understanding natural disease progression \cite{Adams2021RealWorldWearableSensors}. Federated learning frameworks have shown promise for privacy-preserving data collection from wearable devices during daily activities for collaborative model improvement \cite{PATI2024100974}.

Future research could collect long-term data across multiple institutions and establish continuous monitoring protocols that capture both within-day and between-day symptom variability. This approach would also enable the foundation model to adapt to individual patient patterns over time while maintaining cross-patient generalization capabilities for new users. Future research could also explore multi-modal sensor fusion incorporating audio cues and environmental context to enhance detection accuracy in noisy real-world environments.

\subsection{Model Interpretability and Clinical Integration}
Despite our model's strong performance on unseen patients, the transformer-based model operates as a black box, making it challenging for clinicians to understand the decision-making process behind FoG predictions. This lack of interpretability may hinder clinical adoption, as healthcare providers typically require explainable predictions to build trust and make informed treatment decisions. Additionally, integrating the system into existing clinical workflows remains an open challenge.

Future research could focus on explainable AI techniques \cite{Abbas2024FederatedLearningSmartHealthcare} and developing attention visualization methods specifically tailored to IMU-based foundation models. This could
include implementing gradient-based attribution methods to identify which temporal patterns and sensor signals most strongly influence FoG predictions. Additionally, developing standardized interfaces for integrating FoG detection systems with clinical decision support systems would facilitate broader clinical adoption and enable longitudinal tracking of intervention effectiveness. The system's compatibility with standard smartphones makes it accessible to a broader patient population without requiring specialized hardware investments. Integration with existing telehealth platforms could enable remote monitoring and early intervention protocols for patients in rural or underserved areas.
\section{Conclusion}
In this paper, we presented FM-FoG, the first real-time foundation model-based wearable system that achieves FoG detection in previously unseen patients without patient-specific training. Our approach combines self-supervised pretraining on diverse IMU datasets with sensor context integration and an event-triggered activity classifier. Evaluated on 23 PD patients, FM-FoG achieved a 98.5\% F1-score in detection for unseen patients, substantially outperforming existing FoG detection models, time-series FMs, and general-purpose LLMs.

The event-triggered design extended battery life by 43\% while maintaining sub-20ms intervention latency, demonstrating practical deployment feasibility. Our findings show that specialized FMs can outperform larger general-purpose alternatives through domain-specific training. FM-FoG provides a foundation for developing intelligent, adaptive health monitoring technologies that generalize across patients without individual training requirements. 
}

\bibliographystyle{ACM-Reference-Format}
\bibliography{main}

\end{document}